\documentclass[runningheads]{llncs}

\usepackage{eccv}

\usepackage{eccvabbrv}

\usepackage{graphicx}
\usepackage{booktabs}

\usepackage[accsupp]{axessibility}  

\usepackage[pagebackref,breaklinks,colorlinks,citecolor=eccvblue]{hyperref}

\usepackage{orcidlink}

\usepackage{booktabs}
\usepackage{makecell}
\usepackage{amssymb}  
\usepackage{enumitem} 

\usepackage{algorithm}
\usepackage{listings}
\usepackage{multirow}

\newcommand{\bluecell}{\cellcolor{blue!25}}
\newcommand{\redcell}{\cellcolor{red!25}}
\newcommand{\greencell}{\cellcolor{green!25}}

\newcommand{\graycell}{\cellcolor{gray!25}}

\usepackage{xcolor,colortbl}
\definecolor{deemph}{gray}{0.6}

\newcommand*{\Comment}[1]{\hfill\makebox[1em][l]{#1}}%

\usepackage{etoolbox}
\makeatletter
\AfterEndEnvironment{algorithm}{\let\@algcomment\relax}
\AtEndEnvironment{algorithm}{\kern2pt\hrule\relax\vskip3pt\@algcomment}
\let\@algcomment\relax
\newcommand\algcomment[1]{\def\@algcomment{\footnotesize#1}}
\renewcommand\fs@ruled{\def\@fs@cfont{\bfseries}\let\@fs@capt\floatc@ruled
  \def\@fs@pre{\hrule height.8pt depth0pt \kern2pt}%
  \def\@fs@post{}%
  \def\@fs@mid{\kern2pt\hrule\kern2pt}%
  \let\@fs@iftopcapt\iftrue}
\makeatother

\begin{document}

\title{
Fast Encoding and Decoding for Implicit Video Representation
} 

\titlerunning{Fast NeRV}

\author{Hao Chen\inst{1} \and
Saining Xie\inst{2} \and
Ser-Nam Lim\inst{3} \and 
Abhinav Shrivastava\inst{1} }

\authorrunning{Hao Chen et al.}

\institute{
University of Maryland, College Park, USA \and
New York University, USA  \and
University of Central Florida, USA\\ 
\email{\ haochen.umd@gmail.com}}

\maketitle

\begin{abstract}
Despite the abundant availability and content richness for video data, its  high-dimensionality poses challenges for video research. Recent advancements have explored the implicit representation for videos using neural networks, demonstrating strong performance in applications such as video compression and enhancement. However, the prolonged encoding time remains a persistent challenge for video Implicit Neural Representations (INRs).  In this paper,  we focus on improving the speed of video encoding and decoding within implicit representations. We introduce two key components: NeRV-Enc, a transformer-based hyper-network for fast encoding; and NeRV-Dec, a parallel decoder for efficient video loading. NeRV-Enc achieves an impressive speed-up of $\mathbf{10^4\times}$ by eliminating gradient-based optimization. Meanwhile, NeRV-Dec simplifies video decoding, outperforming conventional codecs with a loading speed $\mathbf{11\times}$ faster, and surpassing RAM loading with pre-decoded videos ($\mathbf{2.5\times}$ faster while being $\mathbf{65\times}$ smaller in size).

\end{abstract}

\section{Introduction}
\label{sec:intro}

Video research is a fundamental area in computer vision, owing to its  rich visual content and widespread presence.
However, the immense size of video data presents challenges, as video storage, loading, and processing demands are orders of magnitude larger than those for images.
Recent research has explored the potential of representing high-dimensional video data as deep neural networks~\cite{chen2021nerv,li2022enerv,kim2022scalable,chen2022cnerv,chen2023hnerv,he2023dnerv}. These representations are favored in applications like video compression (achieving up to $1000\times$ size reductions while maintaining good visual quality), and video enhancement~\cite{chen2022vinr,chen2021nerv,kim2022scalable,chen2023hnerv}.
Unlike pixel-wise Implicit Neural Representation (INR) methods that use MLP networks to model individual pixels, the NeRV series~\cite{chen2021nerv,chen2022cnerv,chen2023hnerv,he2023dnerv} employs convolution neural networks to generate entire frames, enhancing video encoding and decoding efficiency.

Despite the efficiency gains achieved by the NeRV series, training a neural network to overfit a given video using gradient-based optimization remains time-consuming. Since video encoding involves mapping input video to NeRV model weights, we propose a straightforward alternative: NeRV-Enc. NeRV-Enc employs a hyper-network to directly generate model weights, thus avoiding the cumbersome encoding process. Our work addresses two fundamental questions: Can a hyper-network be effectively trained on a given video dataset? And can a well-trained hyper-network generalize well to unseen videos?

\begin{figure}[t]
    \centering
    \includegraphics[width=0.99\textwidth]{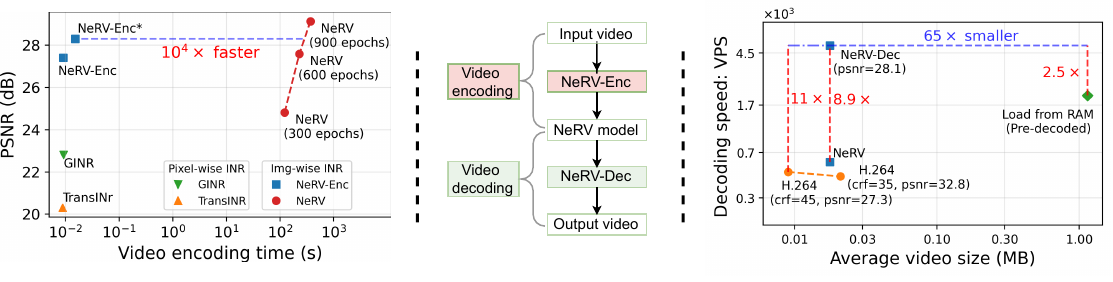}
    \captionof{figure}{
    \textbf{Left:} video encoding for implicit video representations.
    NeRV-Enc is $10^4\times$ faster than NeRV~\cite{chen2021nerv} baseline (with gradient-based optimization).
    * uses a larger encoder and more training videos.
    \textbf{Right:}  video decoding. 
    NeRV-Dec decodes videos $8.9\times$ faster than NeRV and $11\times$ faster than H.264.
    It is even $2.5\times$ faster than loading pre-decoded videos from RAM while being $65\times$ smaller in video size.
    }
    \label{fig:enc-dec-teaser}
    \vspace{-1em}
\end{figure}

Given the outstanding performance of transformer networks in various visual tasks~\cite{dosovitskiy2021an,touvron2021training,liu2021Swin,carion2020end,he2021masked}, we utilize transformers as the hyper-network. This hyper-network takes video patches and initial weight tokens as input, depicted in \cref{fig:hypernerv-archi}. Our research demonstrates that NeRV-Enc provides positive answers to the questions raised earlier. We show that it is indeed possible to train NeRV-Enc on training videos, and it effectively generalizes to unseen videos. In comparison to training NeRV with gradient-based optimization, NeRV-Enc significantly reduces the encoding time, achieving a speed-up of   $\mathbf{10^4 \times}$ (\cref{fig:enc-dec-teaser} Left).

In addition to video encoding, efficient decoding plays a vital role. While a video is encoded just once, it can be decoded countless times. As such, the efficiency of decoding is critical for video playback, streaming, and preview.
In the context of video research, especially in large-scale experiments, video loading involves a more complex decoding pipeline compared to image data loaders. Data loaders that require significant computational resources can prolong research cycles and present substantial hurdles when training video models with extensive datasets~\cite{wu2020multigrid}.
Indeed, data loading significantly hampers training efficiency, causing a slowdown of $46\%$ in video self-supervised learning, as highlighted in \cite{feichtenhofer2022masked}. In this context, video loading stands out as a critical bottleneck, hindering the progress of video research. To address this issue, we present NeRV-Dec, a parallel and efficient video loader based on NeRV.

Traditional video codecs, like H.264, face challenges with complex decoding pipelines, demanding customized design and optimization for specific applications. In contrast, our NeRV-Dec makes vidoe decoding simpler and faster by efficient parallelization.
As shown in \cref{fig:enc-dec-teaser} Right, NeRV-Dec is $8.9\times$ faster than the NeRV baseline and achieves an $11\times$ speed improvement over H.264. Notably, NeRV-Dec outperforms RAM loading of pre-decoded videos by $2.5\times$ faster, while utilizing much less disk storage ($65\times$ smaller). Unlike existing video codecs tailored primarily for CPU use, NeRV-Dec emerges as the superior choice, especially in deep learning research. 
NeRV-Dec efficiently utilizes the power of advanced hardware like GPUs, TPUs, and NPUs, which are already the favored platforms for many users. It achieves this without the need for any specialized design or optimization  for video loading, making it highly compatible and easy to integrate into existing workflows.

We outline the contributions of this paper as follows:
\textbf{a)} We propose the NeRV-Enc architecture for obtaining image-wise implicit video representations (NeRV), achieving a $\mathbf{10^4}$ speedup in encoding compared to conventional gradient-based optimization methods.
\textbf{b)} We extensively explore the architecture design of NeRV-Enc, 
addressing challenges in converting MLP to ConvNets and scaling up the training of NeRV-Enc. 
Specifically, we introduce layer-adaptive weight modulation for NeRV weights, which significantly outperforms previous modulation strategies in terms of efficiency.
\textbf{c)} We introduce NeRV-Dec for parallel decoding of NeRV, achieving an $\mathbf{8.9\times}$ speedup over conventional NeRV decoding. NeRV-Dec also outpaces the common video codec H.264 by $\mathbf{11\times}$ in speed and is $\mathbf{2.5\times}$ faster than loading pre-decoded videos from RAM.
\textbf{d)} Leveraging weight quantization, we reduce video size by  $\mathbf{65\times}$ compared to the original.

\section{Related Work}
\label{sec:related_work}

\noindent\textbf{Implicit Neural Representations.} 
Recent advances in deep learning have given rise to implicit neural representations, which are compact data representations~\cite{chen2021nerv,chen2022cnerv,dupont2021coin,mehta2021modulated}. These representations fit neural networks to signals like images, 3D shapes, and videos. A prominent subset of implicit representations is coordinate-based neural representations, which take pixel coordinates as input and produce corresponding values, such as density or RGB values, using MLP networks. These representations have demonstrated promise in diverse applications, including image reconstruction~\cite{tancik2020fourier, sitzmann2020implicit}, image compression~\cite{dupont2021coin}, continuous spatial super-resolution~\cite{chen2021learning, skorokhodov2021adversarial, anokhin2021image, karras2021alias}, shape regression~\cite{Chen_2019_CVPR, Park_2019_CVPR}, and novel view synthesis for 3D scenes~\cite{mildenhall2020nerf, schwarz2021graf}.

In contrast to coordinate-based methods, NeRV~\cite{chen2021nerv} introduces an image-wise implicit representation that takes the frame index as input and outputs the entire frame without iterative pixel-wise computations. NeRV leverages convolutional neural networks, offering improved efficiency and regression quality compared to coordinate-based methods. By representing videos as neural networks, NeRV transforms video compression into model compression, achieving comparable performance with common video codecs through model pruning, quantization, and entropy encoding. Building on NeRV's success, works~\cite{li2022enerv,chen2022cnerv,chen2023hnerv,he2023dnerv} further enhance efficiency and shows superior performance in video compression, interpolation, and enhancement. Our approach leverages image-wise representations since NeRV series provide higher capacity and faster decoding speed compared to coordinate-based methods.

\noindent\textbf{Hyper-Networks.} 
Hyper-networks~\cite{Ha2017HyperNetworks} are commonly employed to generate model weights for another neural network based on input data or a dataset. The concept of content-adaptive weights is prevalent in deep learning, exemplified in techniques like dynamic convolution~\cite{chen2020dynamic} and conditional convolution~\cite{yang2019condconv}. Various methods have been developed to modulate model weights in a latent space~\cite{Park_2019_CVPR,Mescheder_2019_CVPR,sitzmann2019srns,sitzmann2019siren} rather than generating all weights directly, a strategy that can alleviate learning challenges. In these approaches, the neural network takes both pixel coordinates and a content-adaptive vector for modulation, where the modulated vectors serve as the hyper-networks. TransINR~\cite{chen2022transinr} and GINR~\cite{kim2022generalizable} employ hyper-networks to generate model weights for pixel-wise INRs, suitable for image and video regression.

\noindent\textbf{Efficient Video Codecs.}
Existing video dataloaders rely on traditional codecs like MPEG~\cite{mpeg}, H.264~\cite{wiegand2003overview}, and HEVC~\cite{hevc} to reduce video size. Advanced video codecs like AV1~\cite{chen2018overview} and VVC~\cite{bross2021overview} offer improved compression but at the expense of longer decoding times, hampering data loading speed. Recent developments in deep learning have introduced techniques for video compression, seeking to enhance traditional methods, including image compression, interpolation~\cite{Wu_2018_ECCV,Djelouah_2019_ICCV}, autoencoders~\cite{Habibian_2019_ICCV}, modeling conditional entropy between frames~\cite{liu2020conditional}, and rethinking video compression with deep learning~\cite{Rippel_2019_ICCV,liu2019neural,Agustsson_2020_CVPR}, or refine existing codecs~\cite{khani2021efficient,rippel2021elfvc}. Although these learning-based methods improve bits-distortion performance, they introduce significant decoding latency, rendering them unsuitable for fast video loading.

In contrast, loading videos with implicit representations is straightforward, simple, and fast. NeRV~\cite{chen2021nerv}, a recent implicit video representation, achieves comparable compression performance with traditional video codecs, by reshaping video compression into model compression. The video decoding process in NeRV is a simple feed-forward operation of the convolution network and can be seamlessly deployed across various  devices without specific design or optimization. Drawing inspiration from NeRV, our NeRV-Dec significantly reduces video model size via quantization and entropy encoding. Meanwhile, it improves decoding speed further by enabling scalability and improved parallelization.


\begin{table}[t!]
    \centering
    \begin{tabular}{@{}ll@{}}
        Variable & Definition \\
        \midrule
        $x$  & Input video \\
        $x_t$ & Video frame at time $t$ \\
        $\hat x_t$ & Reconstructed video frame at time $t$ \\
        $g_\phi$ & Hyper-network (w/ parameter $\phi$) \\
        $f_\theta$ & NeRV model (w/ parameter $\theta$) \\
        $\theta_0$ & Initial weight tokens (hyper-network input) \\        
        $\hat \theta'$ & Hyper-network output: video-specific weights \\
        $\theta_1$ & Video-agnostic model weights \\
        $\theta'$ & Final NeRV model weights \\
        $D_\text{train}, D_\text{test}$ & Training and test set \\        
        $C_\text{out}, C_\text{in}$ & Convolution output and input channel width\\
        $K$, $S$ &  Kernel size, upscale factor for NeRV blocks \\
        $M$ & Number of video patches \\
        $N$ & Number of weight tokens \\
        $d$ & Token dimension for transformer encoder \\
        $d_\text{out}$ &  Output dimension for weight tokens \\
        \bottomrule
    \end{tabular}
    \caption{Variables and their definitions.}
    \label{tab:symbol_defination}
    \vspace{-2em}
\end{table}

\section{Method}

To improve the encoding speed of implicit video representation, we present NeRV-Enc. It employs a hyper-network denoted as $g_\phi$ to generate weights $\theta'$ for the NeRV model $f_{\theta}$, based on input video data $x \in \mathbb R^{t \times 3 \times h \times w}$. These learned weights are then used to reconstruct video frames, yielding $\hat{x}_{t} = f_{\theta=\theta'}(t)$. The primary objective is to minimize the reconstruction error between $\hat{x}_t$ and the its ground truth $x_t$ for training videos $D_\text{train}$, while ensuring the generalization to testing videos $D_\text{test}$. 
Additionally, we introduce NeRV-Dec, a method designed for parallel video decoding with good efficiency. 
Please consult \cref{tab:symbol_defination} for a comprehensive list of symbol definitions.

\subsection{Video Encoding: NeRV-Enc}
\label{sec:hypernerv-method}

\begin{figure}[t!]
    \centering
    \includegraphics[width=.65 \linewidth]{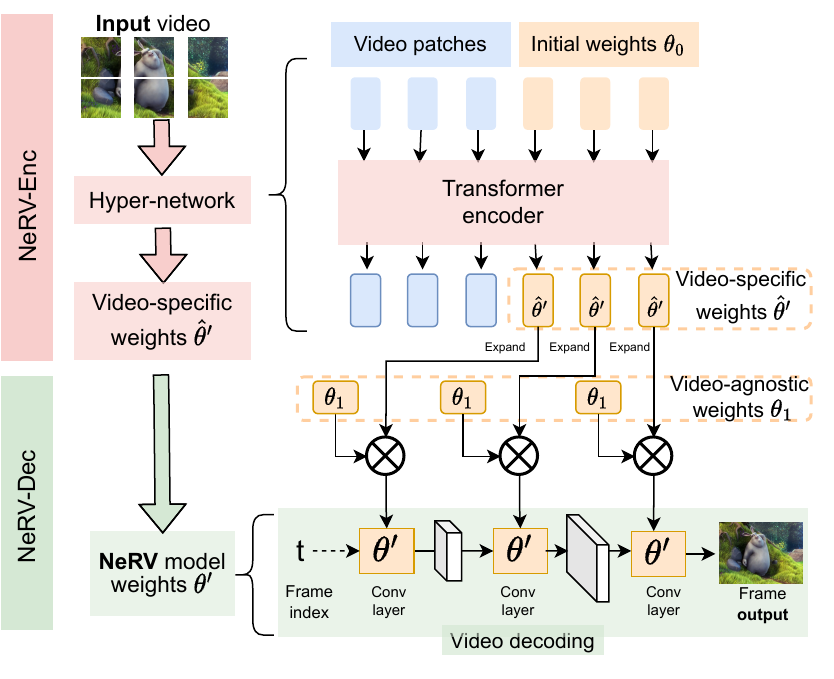}
    \vspace{-1.2em}    
    \caption{
    \textbf{Top:} video encoding. NeRV-Enc processes the input video $x$ to get video-specific weights $\hat \theta'$ using the hyper-network.
    \textbf{Bottom:} video decoding. NeRV-Dec generates final NeRV weights $\theta'$ and reconstruct video $\hat x$.
    }
    \label{fig:hypernerv-archi}
\end{figure}

\noindent\textbf{Generate Video-Specific Weights.}
We employ a Transformer network with $L$ encoder layers as a hyper-network to generate video-specific model weights, denoted as $\hat{\theta}'$. The input video $x$ is partitioned into patches, which are then transformed into patch tokens using a fully connected (FC) layer. Additionally, learned position embeddings are added to these patch tokens. These patch tokens, in conjunction with the initial weight tokens $\theta_0$, form the input tokens. Subsequently, the hyper-network processes these input tokens to produce video-specific weights $\hat{\theta}' \in \mathbb{R}^{d_\text{out} \times N}$, which serves as an compact representation of the video. This encoding process is depicted in \cref{fig:hypernerv-archi} Top and \cref{fig:weights-pipeline-matrix} Left.

\begin{figure}[t!]
    \centering
    \includegraphics[width=.75\linewidth]{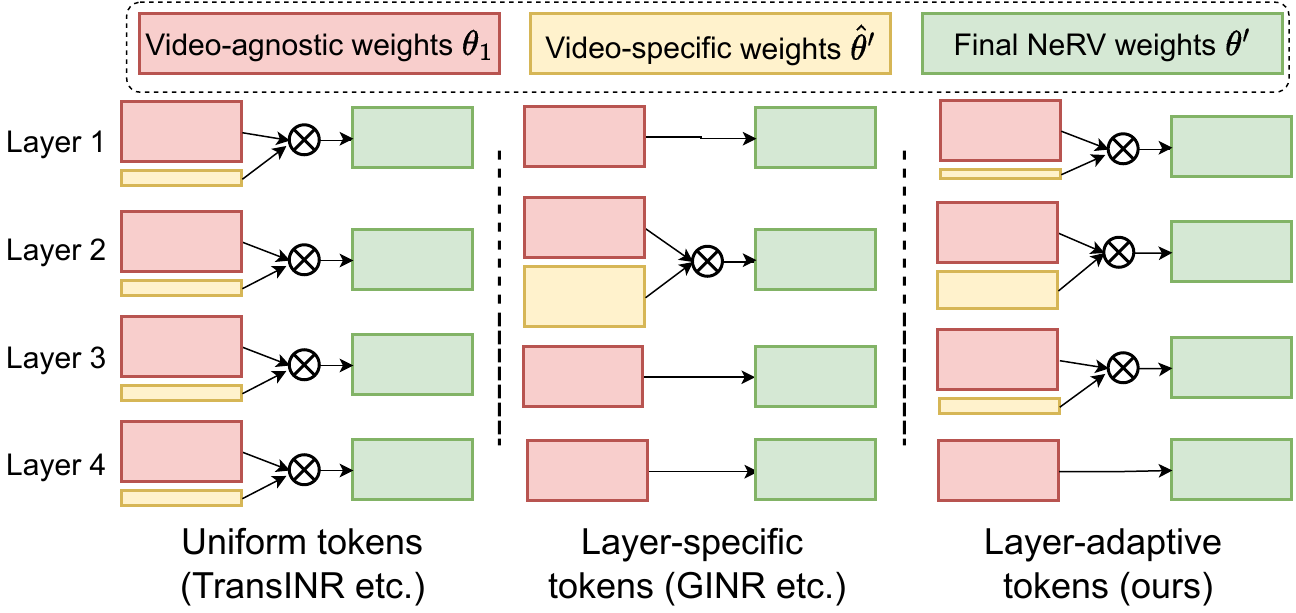}
    \vspace{-.6em}    
    \caption{
    Weight token distributions across layers. 
    \textbf{Left:} Uniform (TransINR~\cite{chen2022transinr}). 
    \textbf{Middle:} Layer-specific (GINR~\cite{kim2022scalable}). 
    \textbf{Right:} Layer-adaptive (ours).
    }
    \label{fig:adaptive-tokens}     

\end{figure}

\vspace{2pt}
\noindent\textbf{Adaptive Weight Tokens across Layers.} 
We've observed that token importance varies across different layers. Unlike TransINR~\cite{chen2022transinr}, which employs an identical number of weight tokens for all layers, or GINR~\cite{kim2022generalizable}, which restricts weight tokens to a specific layer (2nd layer), we introduce a flexible and adaptive approach to weight token distributions. This customization results in three distinct schemes: \textit{uniform weight tokens} (TransINR), \textit{layer-specific weight tokens} (GINR), and \textit{adaptive weight tokens} (our approach), each designed to suit video-specific weight distribution across layers. These diverse weight distributions are visually depicted in \cref{fig:adaptive-tokens}.

\noindent\textbf{NeRV-Enc for Video Restoration.}
Implicit video representations have proven to be good at video restoration tasks such as denoising and inpainting.
Besides reconstruction, we also extend NeRV-Enc to various video restoration tasks. NeRV-Enc takes degraded videos as input and use ground truth videos as supervision. Evaluations on unseen videos indicate that NeRV-Enc effectively addresses these degradation tasks.
Note that while previous implicit methods like NeRV and HNeRV demonstrate competence in restoration tasks, they often require additional supervision, such as masks for inpainting. Furthermore, their training might require reference data for the test video, like high-resolution or de-blurred frames. In contrast, NeRV-Enc leverages large-scale training to learn restoration capabilities and shows robust generalization to unseen videos. This makes NeRV-Enc a more practical and versatile tool for video restoration.

\subsection{Video Decoding: NeRV-Dec}
\label{sec:hypernerv-dec method}

\noindent\textbf{Efficient Video Storage.}
Prior to video decoding, it is necessary to store the video-specific weights efficiently. To achieve storage efficiency, we employ weight quantization and entropy encoding for these weights, similar to compression techniques used in NeRV approaches~\cite{chen2021nerv,chen2023hnerv}.

\begin{figure}[t!]
    \centering
    \includegraphics[width=.75\linewidth]{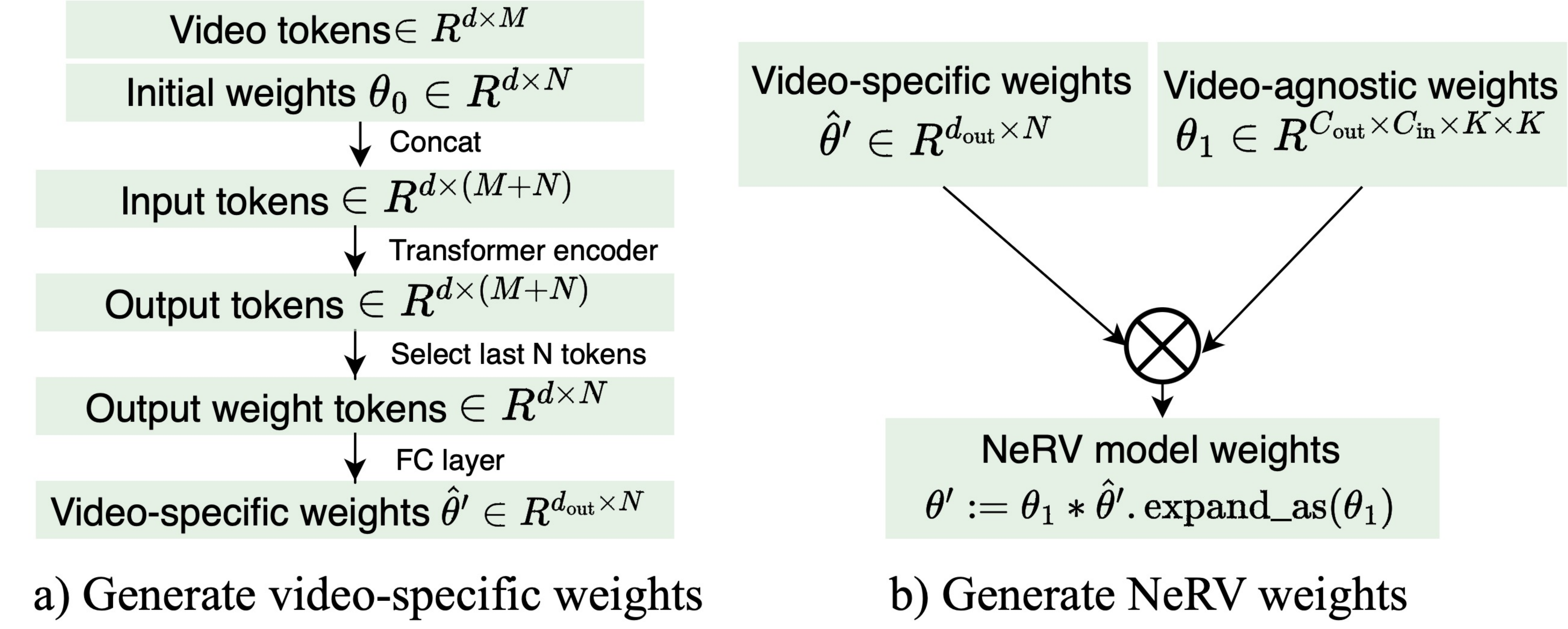}
    \vspace{-.6em}    
    \caption{
    \textbf{Left} Generate video-specific weights $\hat \theta'$  via the hyper-network.
    \textbf{Right} Generate NeRV weights $ \theta'$ by element-wise multiplication of $\hat \theta'$ and video-agnostic weights $\theta_1$.
    }
    \label{fig:weights-pipeline-matrix}

    \vspace{1em}
    \includegraphics[width=.5\linewidth]{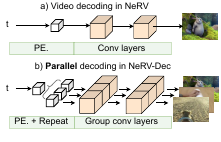}
    \vspace{-1.5em}    
    \caption{
    \textbf{Parallel video decoding} of NeRV-Dec is achieved using group convolution. NeRV decoding is a special case of NeRV-Dec when the group size is 1. 'PE.' denotes position encoding. 'Repeat' indicates  embedding repetition for input expansion.
    }
    \label{fig:nerv-hyperverv-dec}   
    \vspace{-1.5em}
\end{figure}

\noindent\textbf
\noindent\textbf{Generate NeRV Weights.}
After obtaining the video-specific weights $\hat{\theta}'$, we move on to calculate the final weights $\theta'$ for a NeRV model, which we represent as $f_{\theta=\theta'}$. As shown in \cref{fig:weights-pipeline-matrix} (b), the video-specific weights $\hat{\theta}' \in R^{d_\text{out} \times N}$ for a convolution layer (with parameters in $ R^{C_\text{out} \times C_\text{in} \times K \times K } $) might be smaller. To compensate, we introduce \textit{learnable} parameters $\theta_1 \in R^{C_\text{out} \times C_\text{in} \times K \times K }$ for each layer, which are shared across all videos and known as \textit{video-agnostic} parameters. The final weights of the NeRV model, detailed in \cref{fig:weights-pipeline-matrix}, are determined by an element-wise multiplication of the video-agnostic weights $\theta_1$ and the video-specific weights $\hat{\theta}'$.

\vspace{2pt}
\noindent\textbf{Video Decoding Prelimaries for NeRV.} 
With the final NeRV model weights, we access the video frame using a feedforward operation with the frame index $t$. Following a common practice in implicit neural representations~\cite{mildenhall2020nerf, chen2021nerv}, we initially normalize $t$ to the interval [-1, 1], apply positional encoding, and obtain a time embedding vector. This time embedding is used as input to the NeRV model $f_\theta$, which generates the video frame $\hat x_t$. The output of the final layer is adjusted by an output bias of 0.5, considering image normalization within the [0, 1] range. In essence, video decoding can be expressed as $\hat{x}_{t} = f_{\theta=\theta'}(t)$, as shown in \cref{fig:nerv-hyperverv-dec} (a). Efficient video decoding can be accomplished by running NeRV with a batch size encompassing all frames.

\vspace{2pt}
\noindent\textbf{Parallel Decoding for NeRV-Dec.} 
NeRV-Dec enables parallel decoding by running several NeRV models at once, allowing multiple videos to be decoded simultaneously. This process is efficiently executed using group convolution, where each group corresponds to a separate NeRV model, as illustrated in \cref{fig:nerv-hyperverv-dec} (b). Note that NeRV decoding is a special case of NeRV-Dec when video number is 1.
Since video decoding is an essential operation in implicit neural representations, this method of parallelization significantly boosts the training speed of NeRV-Enc. Furthermore, it improves the scalability and efficiency of NeRV-Dec when used as a video data loader.

\subsection{Model Optimization}
\label{sec:optimize-method}

To optimize NeRV-Enc, our objective is to minimize the reconstruction loss between the ground truth frame $x_t$ and the reconstructed frame $\hat{x}_t$, which is expressed as:
\begin{equation}
    \phi^* =  \mathrm{arg\, min}_{\phi,\theta_0,\theta_1} \sum_{x \in D_\text{train}} \sum_t \| f_{\theta=g_{\phi}(x)}(\mathbf t) - \mathbf x_t\|_2^2 \
    \label{equa:train-error}
\end{equation}.
Once we have the optimized NeRV-Enc $\phi^*$, we evaluate it on the test set $D_\text{test}$, addressing the two key questions raised in \cref{sec:intro}: can NeRV-Enc effectively fit training videos, and can it successfully generalize to test videos.

NeRV-Enc consists of three sets of learnable parameters: those for the hyper-network $\phi$, initial weight tokens $\theta_0$, and model-agnostic weights $\theta_1$. For optimization, we utilize a training objective based on the mean square error (MSELoss) between the output frame and the ground truth frames.


\section{Experiment}
\label{sec:experiment}

\subsection{Datasets and Implementation Details}

In our experiments, we utilize three widely-adopted video datasets: Kinetics-400 (K400, our training dataset)\cite{kay2017kinetics}, Something-Something V2 (SthV2)\cite{goyal2017something}, and UCF101~\cite{soomro2012UCF101}. Kinetics-400 comprises about 240k training videos and 20k test videos, each lasting 10 seconds, across 400 classes. Due to the extensive size of the full dataset, we may use a subset of K400 for training, consisting of 25 videos per class. For evaluation, we employ the test sets of K400, SthV2 (about 20k motion-centric videos), and UCF101 (about 3.5k human-centric videos). The quality of the reconstructed videos is evaluated using PSNR and SSIM.

The default video size is $256\times256$ with 8 frames. To preprocess the data, we first resize the input video so that its shorter side is 256. We then perform a center crop to obtain a $256\times256$ clip. Subsequently, we uniformly sample 8 frames from the clip and input them to the model. For data tokenization, we divide the videos into $64\times64$ patches. The video model consists of four NeRV blocks, each with upscale factors of 4, 4, 4, and 4, respectively~\cite{chen2021nerv}. The convolution layers in these blocks maintain a consistent channel width of 16, except for the first block, where the input channel represents the time embedding dimension, and the last block, where the output channel corresponds to the video channels. The kernel size is $1$ for the first convolution layer and $3$ for the subsequent ones.

The default transformer hyper-network is composed of 6 encoder layers with a hidden dimension of 720 and a forward dimension of 2800. We employ the AdamW~\cite{loshchilov2017decoupled} optimizer with a batch size of 32, an initial learning rate of 1e-4.
Our learning rate undergoes a step-wise decay, decreasing by a factor of 0.1 at 90\% of the total training steps. Our implementation is built on PyTorch~\cite{NEURIPS2019_9015}. Model training is conducted using 8 A100 GPUs for all experiments, unless otherwise specified. For video decoding, we test on a machine with 1 A100 GPU and 8 CPUs~\footnote{Intel(R) Xeon(R) Platinum 8259CL CPU @ 2.50GHz}. Additional implementation details and visualization results are available in the supplementary material.

\begin{table*}[t!]
\centering
\resizebox{.98\textwidth}{!}{%
\begin{tabular}{@{}lc|ccc|cc|cccc|cccc}
\toprule
\multirow{2}{*}{Methods} & \multirow{2}{*}{F} & \multirow{2}{*}{\makecell{Encoder \\ size}}  & \multirow{2}{*}{\makecell{ INR \\ size $\downarrow$}}  & \multirow{2}{*}{$\#\hat\theta'$ $\downarrow$} & \multirow{2}{*}{Epoch} & \multirow{2}{*}{\makecell{GPU \\ hrs $\downarrow$}} & \multicolumn{4}{c}{PSNR $\uparrow$} & \multicolumn{4}{c}{SSIM $\uparrow$} \\
 &  &  &  &  &  &  & Train & K400 & SthV2 & UCF101 & Train & K400 & SthV2 & UCF101 \\
\midrule
TransINR~\cite{chen2022transinr} & 4 & 48.0M & 99k & 25k & 150 & 63 & 23.7 & 22.1 & 24.6 & 22.1 & 0.659 & 0.631 & 0.728 & 0.622 \\
GINR~\cite{kim2022scalable} & 4 & 47.6M & 139.4k & 25.6k & 150 & 65 & 24.5 & 23.2 & 25.9 & 23.1 & 0.685 & 0.66 & 0.744 & 0.66 \\
NeRV-Enc & 4 & 47.6M & 85.6k & 24.1k & \redcell{150} & \redcell{9} & \textbf{26.6} & \textbf{26.6} & \textbf{29.4} & \textbf{26} & \textbf{0.756} & \textbf{0.754} & \textbf{0.816} & \textbf{0.752} \\
NeRV-Enc & 4 & 47.6M & 85.6k & 24.1k & \redcell{1000} & \redcell{62} & \bluecell{27.9} & \bluecell{27.5} & \bluecell{30.5} & \bluecell{27.1} & \bluecell{0.794} & \bluecell{0.783} & \bluecell{0.838} & \bluecell{0.783} \\
\midrule
TransINR~\cite{chen2022transinr} & 8 & 48.0M & 99k & 25k & 150 & 119 & 22.3 & 20.3 & 22.8 & 20.7 & 0.626 & 0.595 & 0.703 & 0.591 \\
GINR~\cite{kim2022scalable} & 8 & 47.6M & 139.4k & 25.6k & 150 & 123 & 23.9 & 22.8 & 25.3 & 22.7 & 0.671 & 0.65 & 0.737 & 0.651 \\
NeRV-Enc & 8 & 47.6M & 85.6k & 24.1k & \redcell{150} & \redcell{11} & \textbf{25.8} & \textbf{25.8} & \textbf{28.5} & \textbf{25.2 }& \textbf{0.732} & \textbf{0.727} & \textbf{0.795} & \textbf{0.723 }\\
NeRV-Enc & 8 & 47.6M & 85.6k & 24.1k & \redcell{1500} & \redcell{110} & \bluecell{27.8} & \bluecell{27.4} & \bluecell{30.3} & \bluecell{26.8} & \bluecell{0.791} & \bluecell{0.78} & \bluecell{0.835} & \bluecell{0.778} \\
\midrule
TransINR~\cite{chen2022transinr} & 16 & 48.0M & 99k & 25k & 150 & 234 & 21.5 & 18.4 & 21.1 & 19.2 & 0.615 & 0.555 & 0.678 & 0.561 \\
GINR~\cite{kim2022scalable} & 16 & 47.6M & 139.4k & 25.6k & 150 & 242 & 22.9 & 21.7 & 24.2 & 21.7 & 0.647 & 0.624 & 0.72 & 0.625 \\
NeRV-Enc & 16 & 47.6M & 85.6k & 24.1k & \redcell{150} & \redcell{15} & \textbf{23.6} & \textbf{23.2} & \textbf{25.9} & \textbf{22.9} & \textbf{0.657} & \textbf{0.642} & \textbf{0.731} & \textbf{0.642} \\
NeRV-Enc & 16 & 47.6M & 85.6k & 24.1k & \redcell{2000} & \redcell{200} & \bluecell{25.4} & \bluecell{24.9} & \bluecell{27.7} & \bluecell{24.5} & \bluecell{0.711} & \bluecell{0.693} & \bluecell{0.772} & \bluecell{0.692} \\
\bottomrule
\end{tabular}
}
\caption{
\textbf{NeRV-Enc \vs Pixel-wise INR methods}. 
NeRV-Enc is much faster (up to $15\times$) than pixel-wise methods for training.
 It also shows better quality in reconstructing videos across datasets, as measured by PSNR and SSIM.
`F' refers to frame number, $ \#\hat\theta' $ is the size of video-specific weights.
Training time is measured in `GPU hrs'. 
} 
\label{tab:img-vs-pixel-wise-inr}
\vspace{-1em}
\end{table*}

\subsection{Video Encoding}
\label{sec:encode-experiment}

\noindent\textbf{NeRV-Enc \vs Pixel-wise INRs.} 
We begin by comparing our method to pixel-wise methods TransINR~\cite{chen2022transinr} and GINR~\cite{kim2022generalizable}, which also employ hyper-networks for generating implicit representations.
However, these methods rely on pixel-wise Implicit Neural Representations (INRs), which exhibit inherent inefficiencies, particularly in large-scale training, as demonstrated in \cref{tab:img-vs-pixel-wise-inr}.

\begin{figure*}[t!]
    \centering
    \includegraphics[width=.49\linewidth]{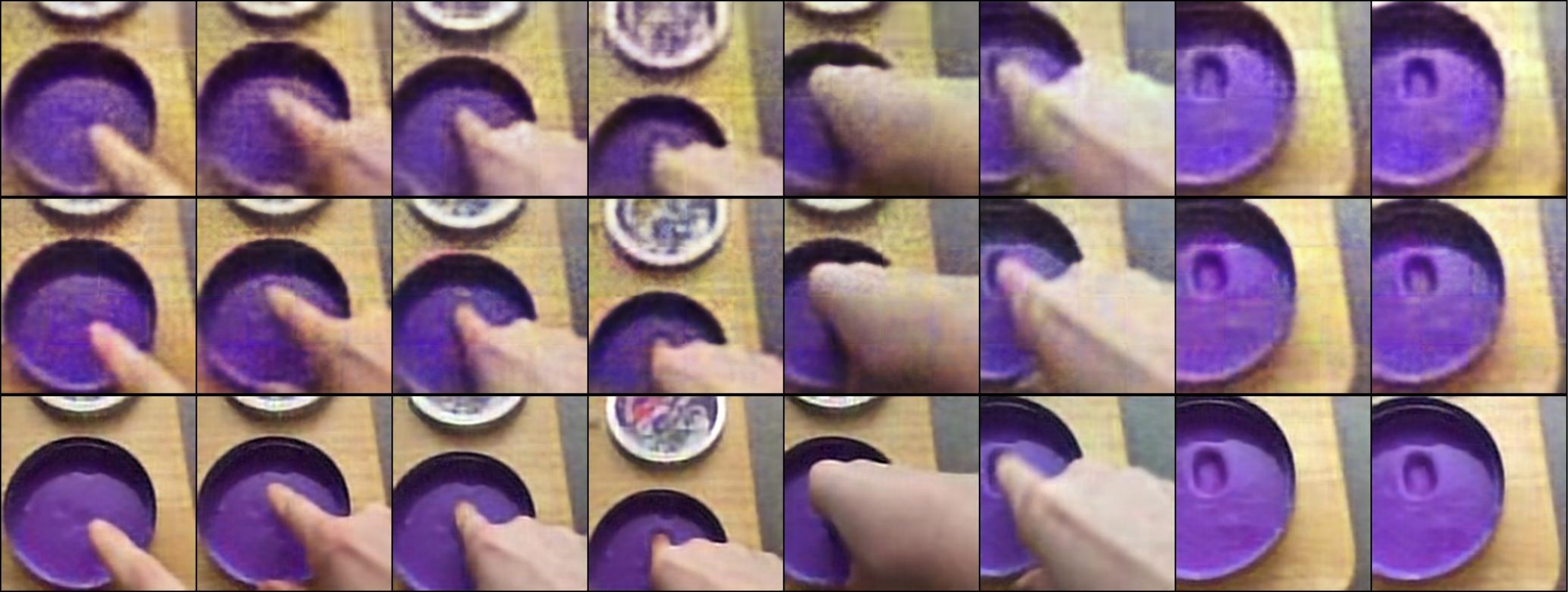}
    \includegraphics[width=.49\linewidth]{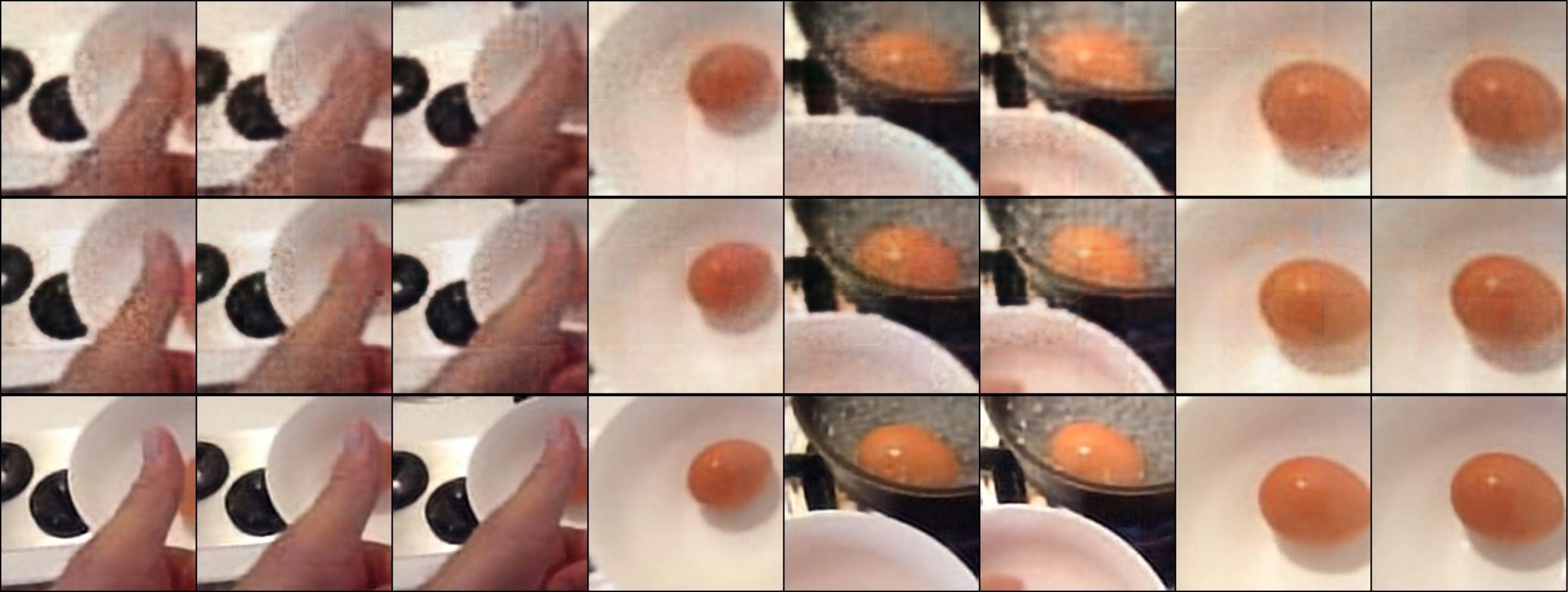}
    \vspace{-.6em}    
    \caption{ 
Visualizations for INR encoding methods: TransINR~\cite{chen2022transinr} (\textbf{Top}), GINR~\cite{kim2022generalizable} (\textbf{Middle}), and NeRV-Enc~(\textbf{Bottom}, ours). Our method excels in reconstructing videos with superior fidelity and fine details.
Best viewed digitally and zoomed in.
    }
    \label{fig:encode-visualization}
\end{figure*}

\textit{Firstly}, pixel-wise INRs are notably slower than image-wise INRs, a fact also demonstrated in NeRV~\cite{chen2021nerv}. For instance, with the same 150-epoch training, NeRV-Enc outperforms pixel-wise methods, achieving $\mathbf{7\times}$ faster training for 4 frames, $\mathbf{10.8\times}$ for 8 frames, and $\mathbf{15.6\times}$ for 16 frames.
\textit{Secondly} despite having a smaller INR model $\theta$ and less model-specific weights $\hat \theta'$ (\ie smaller video size), NeRV-Enc achieves superior video reconstruction quality for both training and testing videos, as measured by PSNR and SSIM metrics. 
The difference in quality becomes more noticeable when comparing pixel-wise methods to NeRV-Enc with \textit{similar training times}. In this comparison, NeRV-Enc outperforms them by \textbf{+4.6, +5.0,} and \textbf{+4.1} PSNR for videos with 8 frames in the K400, SthV2, and UCF101 test sets, respectively.
Qualitative results are provided in \cref{fig:encode-visualization}, comparing NeRV-Enc with pixel-wise methods. These visual comparisons illustrate that NeRV-Enc excels in capturing videos with superior fidelity and fine details.

\begin{table}[t!]
\centering
\renewcommand{\tabcolsep}{2pt}    
\begin{tabular}{@{}ccc|cccc|cccc@{}}
\toprule
\multirow{2}{*}{$\mathcal{E}$}  & \multirow{2}{*}{$\mathcal{S}$}  & \multirow{2}{*}{$\mathcal{N}$} &  \multicolumn{4}{c|}{PSNR $\uparrow$} & \multicolumn{4}{c}{SSIM $\uparrow$} \\
& & & Train & K400 & SthV2 & UCF & Train & K400 & SthV2 & UCF \\
 \midrule
& &  & \graycell{25.8} & \graycell{25.8} & \graycell{28.5} & \graycell{25.2} & \graycell{0.732} & \graycell{0.727} & \graycell{0.795} & \graycell{0.723} \\
\midrule
\checkmark & & & 27.8 & 27.4 & 30.4 & 26.9 & 0.791 & 0.782 & 0.837 & 0.781 \\
& \checkmark &  & 26.5 & 26.7 & 29.5 & 26.2 & 0.756 & 0.762 & 0.822 & 0.759 \\
& & \checkmark & 27.8 & 27.9 & 30.9 & 27.4 & 0.792 & 0.799 & 0.852 & 0.802 \\
\midrule
\checkmark & \checkmark & \checkmark  & \bluecell{28.1} & \bluecell{28.4} & \bluecell{31.6} & \bluecell{28.1} & \bluecell{0.803} & \bluecell{0.808} & \bluecell{0.862} & \bluecell{0.817} \\
\bottomrule
\end{tabular}
\caption{
\textbf{Scale NeRV-Enc} by increasing training epochs $\mathcal{E}$, hyper-nerwork size $\mathcal{S}$, training video number $\mathcal{N}$.
}
\label{tab:encode-improvement}
\end{table}

\vspace{2pt}
\noindent\textbf{Scale NeRV-Enc.}
We also investigate scaling techniques to further improve the performance of NeRV-Enc: longer training epochs, more training videos, and a larger hyper-network. The results are presented in \cref{tab:encode-improvement}.  It is evident that expanding the training epochs and incorporating more training videos leads to substantial improvements in reconstruction quality, with a notable increase of $+2$ in PSNR and $+0.06$ in SSIM for both training and testing videos. Additionally, increasing the size of the hyper-network, together with a larger dropout ratio for transformer layers as well, enhances reconstruction performance. 
The integration of these three techniques results in our final NeRV-Enc model, leading to \textbf{+2.6, +3.1}, and \textbf{+2.9} PSNR improvements for three test sets. 
We find that NeRV-Enc's reconstruction performance improves with increased computational resources and provide more ablation results in the appendix.

\begin{figure}[t!]
    \centering
    \includegraphics[width=.65\linewidth]{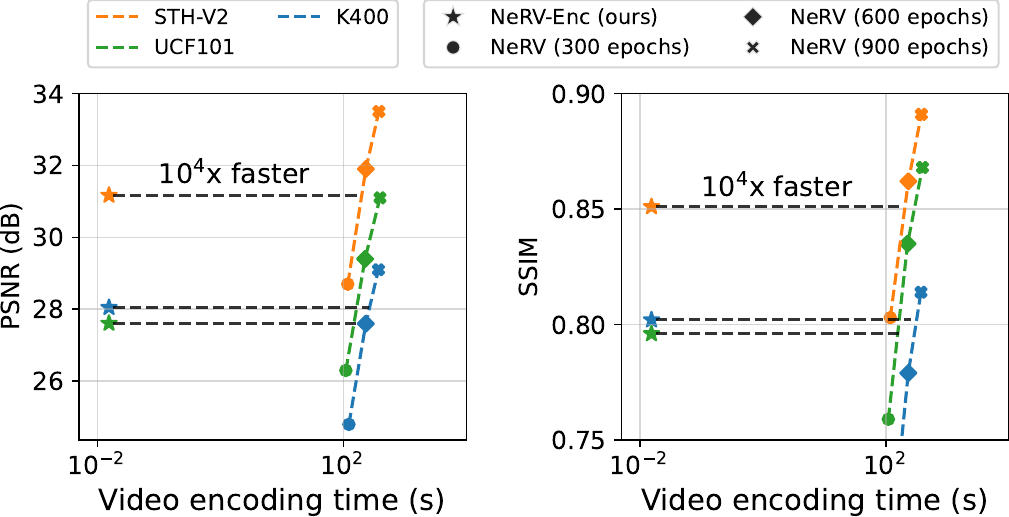}
    \vspace{-.6em}
    \caption{ 
    \textbf{Encoding speed.} NeRV-Enc (ours) is $10^4\times$ faster than NeRV~\cite{chen2021nerv} baseline (using gradient-based optimization) across multiple datasets.
    }
    \label{fig:encode-main}
\end{figure}

\vspace{2pt}
\noindent\textbf{NeRV-Enc \vs NeRV.}
Using the finalized NeRV-Enc, we compare it with the NeRV baseline~\cite{chen2021nerv} which uses gradient-based optimization for model fitting and video encoding. The results, depicted in \cref{fig:encode-main}, highlight the effective generalization of NeRV-Enc across various video datasets, including K400, SthV2 and UCF101. 
Significantly, NeRV-Enc demonstrates a remarkable encoding acceleration, being $\mathbf{10^4\times}$ times faster than the NeRV baseline, yet maintaining comparable output quality, as measured by PSNR and SSIM metrics. It's noteworthy that NeRV-Enc enables \textit{real-time video encoding}, positioning implicit video representation as a viable and efficient option for video codec.

\begin{figure}[t!]
    \centering
    \includegraphics[width=.32\linewidth]{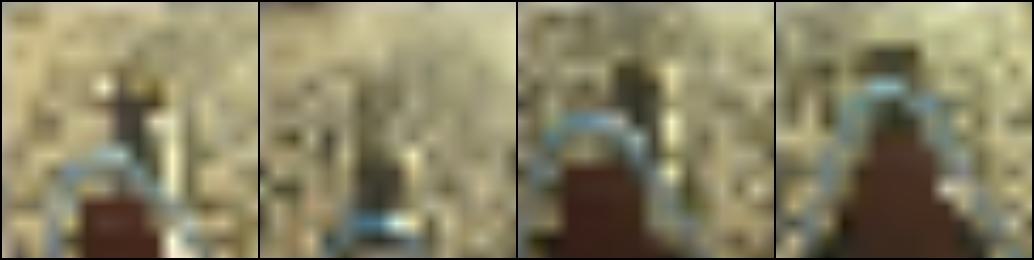}
    \includegraphics[width=.32\linewidth]{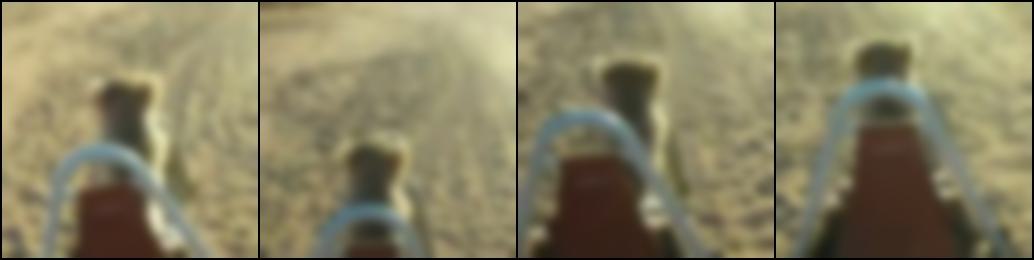}
    \includegraphics[width=.32\linewidth]{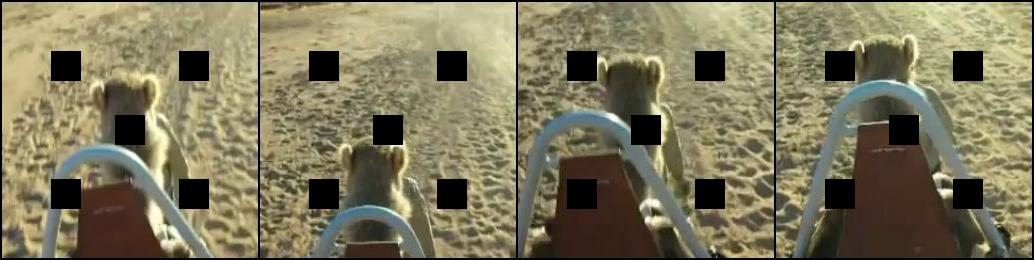}
    \includegraphics[width=.322\linewidth]{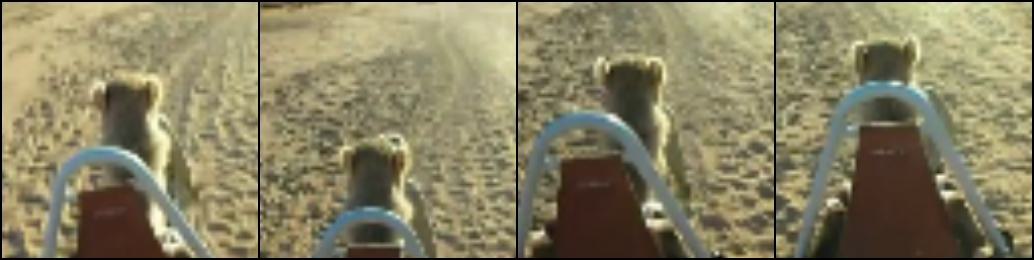}
    \includegraphics[width=.322\linewidth]{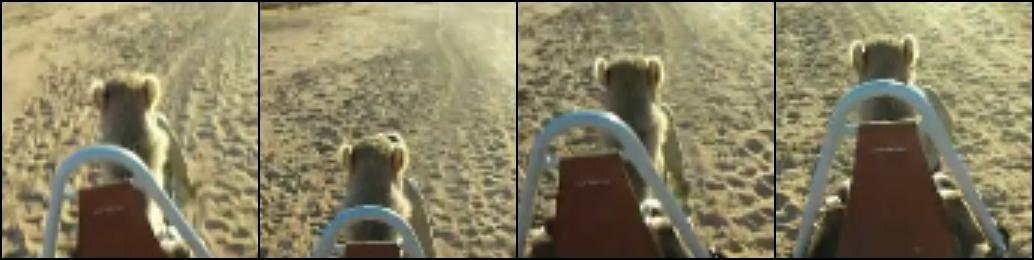}
    \includegraphics[width=.322\linewidth]{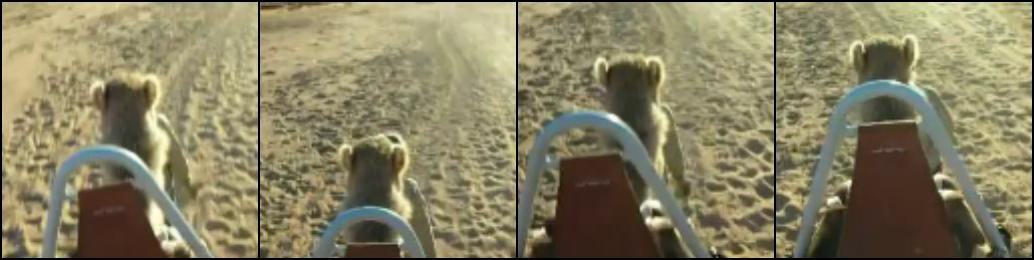}    
\vspace{-0.8em}
    \caption{ 
\textbf{Top:} input videos with various degradations from test set.
\textbf{Bottom:} output videos of NeRV-Enc. 
\textit{Left}: downsampled; \textit{Middle}: blurred; \textit{Right}: mask.
    }
    \label{fig:restore}    
\end{figure}

\vspace{2pt}
\noindent\textbf{NeRV-Enc for Video Restoration.}
Like prior methods that use implicit representations for videos, we found that NeRV-Enc  is also effective for video restoration tasks. Our findings, detailed in \cref{fig:restore}, cover three types of video degradation: downsampling, blurring, and masking of input videos. 
We observed that these \textit{degradations in pixel space are effectively restored in implicit space}. While our primary focus in this paper is on fast encoding and decoding, we have included quantitative results in the appendix for further reference.

\begin{table}[t!]
\centering
\resizebox{.98\linewidth}{!}{%
\begin{tabular}{llccccc}
\toprule
Methods & \textbf{Layer 1/2/3/4 params} & \textbf{TotalParam} & \textbf{Train} & \textbf{K400} & \textbf{SthV2} & \textbf{UCF101} \\
\midrule
Layer-uniform~\cite{chen2022transinr} & 4.1k\_9.2k\_9.2k\_6.9k & 29.4k & 22.1 & 21.5 & 24.1 & 21.6 \\
Layer-specific~\cite{kim2022scalable} & 0\_36.9k\_0\_0                & 36.9k                & 25.5           & 25.4          & 28.1            & 24.8            \\
\midrule
\multirow{5}{*}{\makecell{Layer-adaptive \\ (ours)} } & 0\_18.4k\_0\_0                & 18.4k                & 25.1           & 25.0          & 27.7            & 24.4            \\
& \graycell{1.1k\_18.4k\_4.6k\_0 }         & \graycell{\textbf{24.1k}}                & \graycell{\textbf{25.8}}  & \graycell{ \textbf{25.8}}          & \graycell{ \textbf{28.5}}            & \graycell{ 25.2}            \\
& 2.4k\_18.4k\_4.6k\_0          & 25.4k                & 25.5           & 25.4          & 28.2            & 25.0            \\
& 1.1k\_18.4k\_9.2k\_0          & 28.7k                & 25.8           & 25.8          & 28.5            & \textbf{25.3 }           \\
& 1.1k\_18.4k\_4.6k\_0.1k      & 24.2k    & 25.7 & 25.6 & 28.3 & 25.1 \\
\bottomrule
\end{tabular}
}
\caption{\textbf{Ablation Study} on layer-daptive modulation. We compare uniform tokens (TransINR~\cite{chen2022transinr}), layer-specific tokens (GINR~\cite{kim2022scalable}), and our layer-adaptive tokens. Our approach surpasses the layer-specific method in reconstruction performance (PSNR \textuparrow) and achieves a 50\% reduction in video size (total parameters \textdownarrow).}
\label{tab:layer-wise-ablation}
\vspace{-2em}
\end{table}

\noindent\textbf{Ablation for Layer-adaptive Modulation.}
The ablation study on layer-wise modulation is presented in \cref{tab:layer-wise-ablation}, demonstrating that even with reduced parameters in Layer 2 (see row 3), we can achieve results on par with the layer-specific approach (see row 2).
 Moreover, we incrementally increase parameters in other layers (rows 4 to 7) until no additional improvements were observed.
 In examining the distribution of weight tokens, we evaluate uniform tokens (TransINR~\cite{chen2022transinr}), layer-specific tokens (GINR~\cite{kim2022scalable}), and our proposed layer-adaptive weight tokens, which use 29.4K, 36.9K, and 24.1K video-specific weights, respectively. The results indicate that our approach not only achieves better reconstruction quality but also does so with fewer video-specific parameters, contributing to reduced video size and enhanced compression efficiency.
Note that our layer-adaptive modulation (highlighted in the gray row) not only surpasses the performance of the layer-specific method but also achieves a  \textit{50\% reduction} in the total number of parameters (video size).

\begin{figure}[t!]
    \centering
    \includegraphics[width=.75\linewidth]{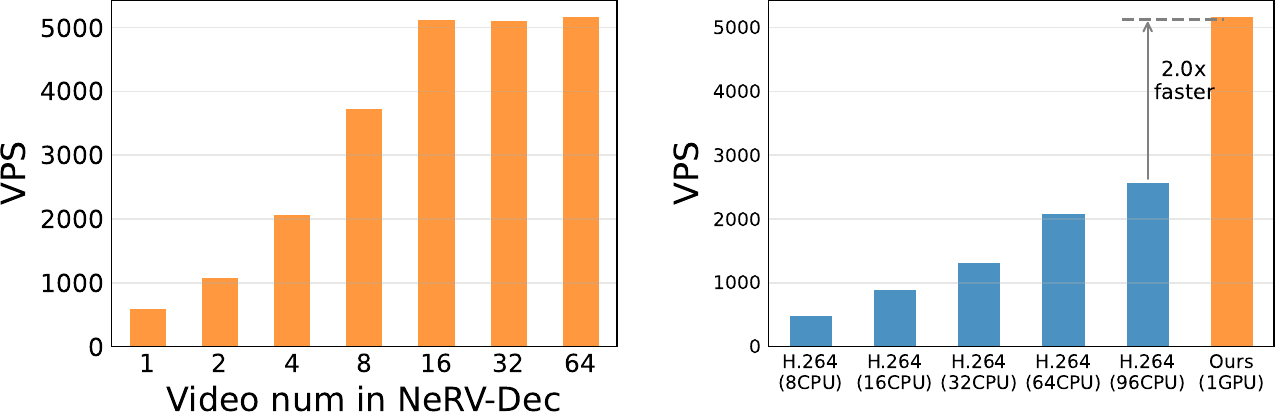}
    \caption{ 
    \textbf{Left:} NeRV-Dec's decoding speed scales efficiently with increasing video numbers. Note NeRV decoding is a special case of NeRV-Dec when video number is 1 (the leftmost bar).
    \textbf{Right:} NeRV-Dec is $\mathbf{2\times}$ faster than H.264 with $\mathbf{96}$  CPUs.
    }
    \label{fig:nerv-dec-h.264-load}
\end{figure}

\subsection{Video Decoding}

\noindent\textbf{Parallelization.}
NeRV-Dec improves upon NeRV by using multiple stacked models to decode several videos at once. 
Its efficiency comes from a shared time embedding layer and a group convolution process that can be parallelized. We evaluate this parallelization by varying the number of videos for decoding. NeRV decoding is a specific case of NeRV-Dec when decoding video number is 1. As shown in \cref{fig:nerv-dec-h.264-load} (Left), there is a near-linear speedup in decoding as video batch size increases from 1 to 16. Beyond a batch size of 16, decoding speed plateaus around 5200 videos per second, maximizing GPU usage.
Video decoding (data loading) is crucial as videos are decoded repeatedly, essential for playback, streaming, and research. Preview of multiple videos is also common for video platforms. Typically, videos are divided into clips or picture groups for storage. Our parallel decoding significantly boosts video loading efficiency.

\textit{Comparison with H.264.}
NeRV-Dec outperforms a traditional data loader using the H.264 codec (485 videos per second for 8 CPUs), achieving a $11\times$ increase in loading speed.
To further understand the decoding performance of H.264, we explore various ways for its speed improvement. Surprisingly, transitioning from CPU to GPU as the decoding device does not yield speed gains, except for larger videos. Consequently, we increase the number of CPUs and data loader workers, presenting the results in \cref{fig:nerv-dec-h.264-load} (Right). The results show that while augmenting the CPU count enhances video loading speed, the improvements diminish, especially beyond 16 CPUs. It's worth noting that even with $\mathbf{96}$ CPUs, H.264 is still $\mathbf{2\times}$ slower than our NeRV-Dec. Given that powerful chips like GPUs are already preferred for deep learning practitioners, the video loading advantages of NeRV-Dec can be leveraged without the need for additional hardware or specialized design and optimization.

\begin{table*}[t!]
\centering
\renewcommand{\tabcolsep}{3pt}    
\resizebox{.98\linewidth}{!}{%
\begin{tabular}{l|c|c|ccc|ccccc}
\toprule
\multirow{2}{*}{} & \multirow{2}{*}{RAM} & \multirow{2}{*}{\makecell{AV1 \\ CRF 60}} & \multicolumn{3}{c|}{H.264} &  \multicolumn{5}{c}{NeRV-Dec (ours)} \\ 
 &  &  & CRF 35 & CRF 40 & CRF 45 & 8 bits & 7 bits & \redcell{6 bits} & 5 bits & 4 bits \\
\midrule
Size \textdownarrow & \greencell{1.15MB} & 21.9KB & 20.4KB & 13.1KB & \bluecell{\textbf{8.7KB}} & 23.7KB & 20.7KB & \redcell{17.7KB} & 14.7KB & 11.6KB \\
PSNR \textuparrow & - & 32.4 & \bluecell{\textbf{32.8}} & 30.0 & 27.3 & 28.4 & 28.3 & \redcell{28.1} & 27.5 & 25.6 \\
SSIM \textuparrow  & - & 0.910 & \bluecell{\textbf{0.912}} & 0.860 & 0.788 & 0.808 & 0.807 & \redcell{0.802} & 0.784 & 0.712 \\
VPS \textuparrow  & 2031 & 313 & 447 & 460 & 485 & \multicolumn{5}{c} {\redcell{\textbf{5175}}} \\
\bottomrule
\end{tabular}
}
\caption{
\textbf{Detailed comparison.} 
NeRV-Dec reduces video size by $65\times$  via weight quantization, while being $2.5\times$ faster than loading pre-decoded videos from RAM.
Although AV1 and H.264 provide better compression (smaller video size) and video quality (higher PSNR and SSIM), NeRV-Dec decodes videos much faster (higher VPS).
}
\label{tab:nerv-dec-size-quality-speed}
\end{table*}

\begin{figure*}[t!]
    \centering
    \includegraphics[width=.49\linewidth]{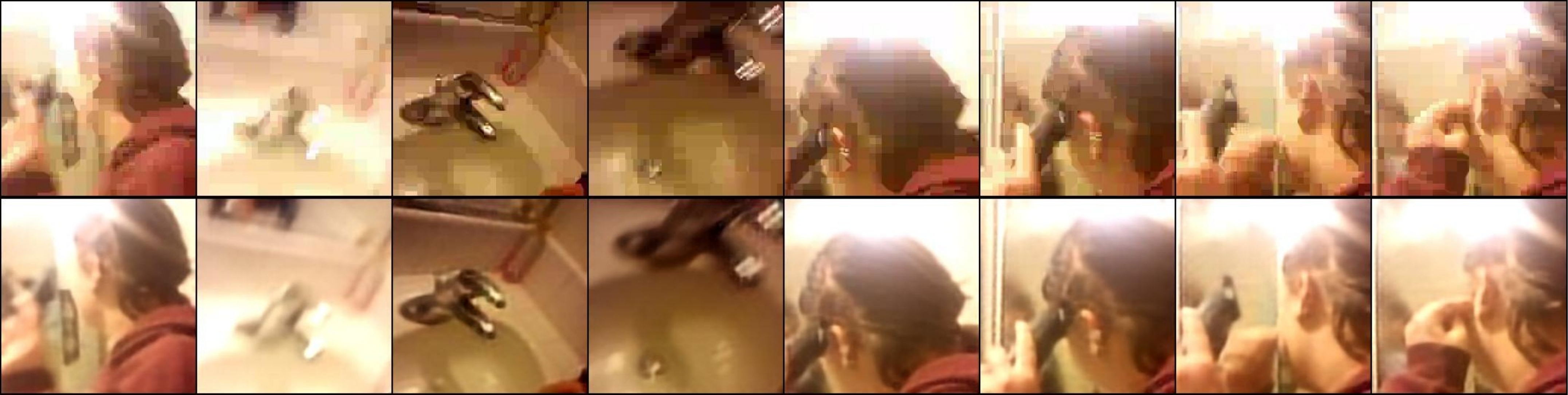}
    \includegraphics[width=.49\linewidth]{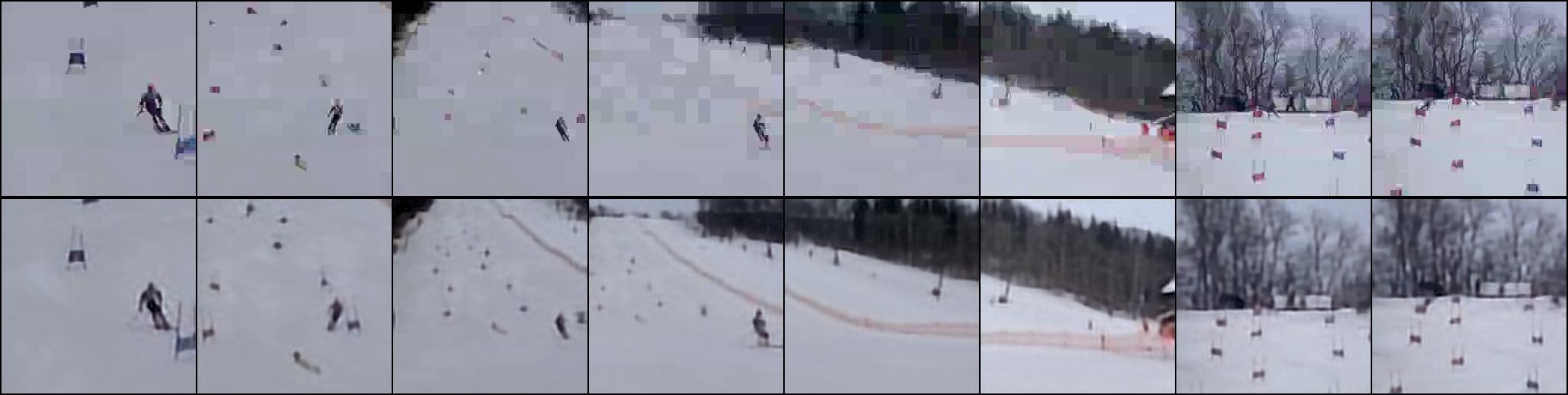}

    \caption{ 
    \textbf{Visual comparison:} H.264 (\textbf{Top}) \vs NeRV-Dec (\textbf{Bottom}) at similar PSNR. H.264 exhibits noticeable blocking artifacts, whereas NeRV-Dec provides a more visually appealing result.
    Best viewed digitally and zoomed in.
    }
    \label{fig:decode-visualization}    
\end{figure*}

\vspace{2pt}
\noindent\textbf{Detailed Comparisons.}
Besides the video decoding speed, we conduct a comprehensive assessment of NeRV-Dec, examining video reconstruction quality and compression performance. Our method undergoes comparisons against various video representations, including traditional codecs like H.264 and AV1, as well as RAM reading (load pre-decoded videos from RAM) for  efficient data loading. Detailed results are presented in \cref{tab:nerv-dec-size-quality-speed}.

\textit{Video Compression.}
To compress the video size in NeRV-Dec, we implement quantization to the video-specific weights ($\hat \theta'$) and employ Huffman encoding. These techniques substantially decreases disk storage requirements while maintaining video quality. It retains 99\% of the PSNR and SSIM values of the original model using just \textit{6 bits} per parameter. 
With a reduced video size (\textbf{$65\times$} smaller), NeRV-Dec even surpasses loading videos from RAM, achieving a \textbf{$2.5\times$} faster speed with GPU assistance. As deep learning platforms commonly utilize powerful hardware like GPUs for computation, the fast and simple video loading provided by NeRV-Dec can significantly enhance video research and alleviate data loading bottlenecks.

\textit{Comparison with Traditional Codecs.}
While the compression ratio of our method is slightly lower than that of conventional codecs such as H.264 and AV1, our method is $11\times$ faster in video decoding and can be easily implemented on most devices without any specific design or optimization. As outlined in \cref{tab:encode-improvement}, enhancing the scale of NeRV-Enc's training by employing additional resources (like more videos for training, extended training epoches, or a larger hyper-network) could yield further improvements and lead to better compression performance.
Beyond just numerical comparisons, we provide visual comparisons between H.264 and NeRV-Dec at an equivalent PSNR of 28.1, as shown in \cref{fig:decode-visualization}. These visual examples clearly demonstrate the blocking artifacts present in H.264, emphasizing the superior visual quality offered by NeRV-Dec. This This advantage highlights the potential preference for NeRV-Dec in scenarios where visual quality is a priority.

\subsection{Limitations and Future Works}
\textit{Hybrid INRs} enhance the NeRV framework by utilizing inputs that are either content-adaptive embeddings or learnable 2D/3D grids. These enhancements allow for superior performance in tasks like video compression or interpolation compared to the original NeRV. We extend NeRV-Enc to also produce these video-specific embeddings or learnable grid features, which then serve as inputs for NeRV.
We show results in \cref{tab:inr-type} and NeRV-Enc is adaptable to diverse INR methods. While NeRV-Enc shows promise in facilitating hybrid INRs, further investigation is required to  achieve better results. The hybrid INR approach (HNeRV), which involves learning both input embedding and decoder weights, necessitates a more complex design.

NeRV-Enc does not yet match the performance of established video codecs such as H.264 in video compression. Closing this performance gap is another focus for our future work. Meanwhile, further exploration of reconstruction loss and restoration tasks  holds promise to produce visually-appealing results.

\begin{table}[t!]
\centering
\begin{tabular}{@{}c|cccc|cccc@{}}
\toprule
& \multicolumn{4}{c}{PSNR $\uparrow$} & \multicolumn{4}{c}{SSIM $\uparrow$} \\
 INR & Train & K400 & SthV2 & UCF101 & Train & K400 & SthV2 & UCF101 \\
 \midrule
NeRV & 25.8 & 25.8 & 28.5 & 25.2 & 0.732 & 0.727 & 0.795 & 0.723 \\
HNeRV & 23.1 & 22.9 & 25.1 & 22.6 & 0.647 & 0.644 & 0.727 & 0.642 \\
\bottomrule
\end{tabular}
\caption{NeRV-Enc results for NeRV and hybrid INR (HNeRV).}
\label{tab:inr-type}
\vspace{-2em}
\end{table}

\section{Conclusion}
In this paper, we introduce NeRV-Enc, a hyper-network that improve encoding speed by generating weights for the NeRV model. Our findings reveal that NeRV-Enc significantly accelerates the encoding process, achieving a remarkable speed-up of $10^4$ times compared to the traditional training of NeRV using gradient-based optimization. 
Additionally, we present NeRV-Dec, a parallel video decoder that surpasses traditional codecs in speed by $11\times$, and outperforms the speed of loading pre-decoded videos from RAM  by $2.5\times$.

%
%
\bibliographystyle{splncs04}
\bibliography{main}

\clearpage
\setcounter{page}{1}

\appendix
\section{Implementation Details}

\subsection{Pseudocode of NeRV-Enc and NeRV-Dec}
We firstly provide pseudocode for NeRV-Enc and NeRV-Dec in \cref{alg:code}.


\begin{algorithm}[h!]
\caption{Pseudocode in a PyTorch style.}
\label{alg:code}
\algcomment{\fontsize{7.2pt}{0em}\selectfont 
\texttt{FC}: fully connected layer; \\
\texttt{MSELoss}: mean square error loss.
\vspace{-1.em}
}
\definecolor{codeblue}{rgb}{0.25,0.5,0.5}
\lstset{
    mathescape=true,
  backgroundcolor=\color{white},
  basicstyle=\fontsize{7.2pt}{7.2pt}\ttfamily\selectfont,
  columns=fullflexible,
  breaklines=true,
  captionpos=b,
  commentstyle=\fontsize{7.2pt}{7.2pt}\color{codeblue},
  keywordstyle=\fontsize{7.2pt}{7.2pt},
  escapechar=\&
}
\begin{lstlisting}[language=python]
############   1) Video encoding: NeRV-Enc ############
# Input: video $x$, initial weights $\theta_0$
# Output:  video-specific weights ${\hat \theta}'$  

# Video tokenization
x = FC$_1$(x.tokennize())           &\Comment{} & # d \times M

# Concat video patches and initial weights as input
x = x.concat($\theta_0$)     &\Comment{} & # d \times (M+N)

# Hypernetwork $g_\phi$ output video-specific weights $\hat{\theta}'$
$\hat{\theta}'$  = g_$\phi$.forward(x)[-N:].  &\Comment{} & # d \times N
$\hat{\theta}'$ = FC$_2$($\hat{\theta}'$)   &\Comment{} &  # C$_\text{out}$ \times N

############ 2) Video decoding: NeRV-Dec ############
$\hat{\theta}'$ = $\hat{\theta}'$.expand_as($\theta_1$)    &\Comment{} &  # broadcast into needed shape
$\theta'$ = $\theta_1$ * $\hat{\theta}'$  &\Comment{} &  # C$_\text{out}$ \times C$_\text{in}$ \times K \times K

# Initial NeRV model $f_\theta$ with generated weights
f_$\theta$.reset_parameter($\theta'$)  

# Input frame index $t$, and output video frame $\hat x$$_\text{t}$
$\hat{x}_t$ = f_$\theta$.forward(t)

###############  3) Model optimization ###############
# Compute loss and backward gradients
loss = MSELoss($\hat{x}_t$, $x_t$)
loss.backward()

# update all learnable parameters
update([$\phi$, $\theta_0$, $\theta_1$])

\end{lstlisting}
\end{algorithm}

\begin{table}[t!]
\centering
\renewcommand{\tabcolsep}{3pt}    
\begin{tabular}{@{}c|cccc|cccc@{}}
& \multicolumn{4}{c}{PSNR $\uparrow$} & \multicolumn{4}{c}{SSIM $\uparrow$} \\
 & Train & K400 & SthV2 & UCF101 & Train & K400 & SthV2 & UCF101 \\
 \midrule
\multicolumn{9}{l}{\textbf{Epoch} ablation} \\
150 & \graycell{25.8} & \graycell{25.8} & \graycell{28.5} & \graycell{25.2} & \graycell{0.732} & \graycell{0.727} & \graycell{0.795} & \graycell{0.723} \\
300 & 26.8 & 26.7 & 29.5 & 26.2 & 0.763 & 0.757 & 0.819 & 0.757 \\
600 & 27.2 & 27.1 & 30 & 26.6 & 0.774 & 0.768 & 0.827 & 0.766 \\
1200 & 27.6 & 27.3 & 30.2 & 26.7 & 0.787 & 0.776 & 0.832 & 0.774 \\
1800 & \bluecell{27.8} & \bluecell{27.4} & \bluecell{30.4} & \bluecell{26.9} & \bluecell{0.791} & \bluecell{0.782} & \bluecell{0.837} & \bluecell{0.781} \\

 \midrule
 \multicolumn{9}{l}{\textbf{Encoder size} ablation} \\
47.6M & \graycell{25.8} & \graycell{25.8} & \graycell{28.5} & \graycell{25.2} & \graycell{0.732} & \graycell{0.727} & \graycell{0.795} & \graycell{0.723} \\
125M & 26.4 & 26.2 & 28.9 & 25.6 & 0.751 & 0.743 & 0.806 & 0.737 \\
251M & \bluecell{26.5} & \bluecell{26.7} & \bluecell{29.5} & \bluecell{26.2} & \bluecell{0.756} & \bluecell{0.762} & \bluecell{0.822} & \bluecell{0.759} \\

404M & 26.5 & 26.5 & 29.3 & 25.9 & 0.753 & 0.755 & 0.816 & 0.751  \\
 \midrule
  \multicolumn{9}{l}{\textbf{Video number} ablation} \\
10k & \graycell{25.8} & \graycell{25.8} & \graycell{28.5} & \graycell{25.2} & \graycell{0.732} & \graycell{0.727} & \graycell{0.795} & \graycell{0.723} \\
20k & 26.7 & 26.6 & 29.4 & 26 & 0.757 & 0.752 & 0.814 & 0.751 \\
40k & 27 & 26.9 & 29.7 & 26.4 & 0.764 & 0.761 & 0.821 & 0.762 \\
80k & 27.6 & 27.7 & 30.6 & 27.2 & 0.791 & 0.791 & 0.845 & 0.794 \\
240k & \bluecell{27.8} & \bluecell{27.9} & \bluecell{30.9} & \bluecell{27.4} & \bluecell{0.792} & \bluecell{0.799} & \bluecell{0.852} & \bluecell{0.802} \\
\bottomrule
\end{tabular}
\caption{
\textbf{NeRV-Enc ablations}. For ablation of weight tokens, we compare uniform tokens (TransINR~\cite{chen2022transinr}), layer-specific tokens (GINR~\cite{kim2022scalable} at the 2nd layer), and our proposed layer-adaptive weight tokens. Please refer to the main paper for their distinction.
}
\label{tab-appendix:ablation}
\end{table}

\subsection{Scaling the training of NeRV-Enc}
We explore factors such as the number of training videos, training epochs, encoder size, and weight token distributions, as outlined in \cref{tab-appendix:ablation}. Generally, NeRV-Enc's reconstruction performance improves with increased computational resources, although the gains tend to plateau as training duration becomes adequate. Higher dropout ratios are essential for achieving improved generalization in longer training epochs or with larger hyper-networks.
These ablations underscore that the reconstruction quality of NeRV-Enc can be significantly enhanced by scaling the training with additional resources, including more training videos, longer training epochs, and larger hyper-networks.

\subsection{NeRV-Enc for video restoration tasks.}
Our NeRV-Enc framework is versatile across various downstream tasks, and shows robust restoration quality for various degradations.
Results in \cref{fig-appendix:restore} and \cref{tab:restore} demonstrate that the reconstruction quality for downsampled, blurred, and masked input videos is comparable to that achieved through conventional video regression.
\textit{This underscores the framework's effectiveness in restoring common pixel degradations within the implicit space.
}


\begin{table}[h]
\centering
\begin{tabular}{l|cccc|cccc}
\multirow{2}{*}{\makecell{Input \\ degradation}} & \multicolumn{4}{c|}{Input PSNR} & \multicolumn{4}{c}{Output PSNR} \\ 
 & Train & k400 & sth-v2 & ucf101 & Train & k400 & sth-v2 & ucf101 \\ 
\midrule
Downsample & 20.1 & 20.3 & 22.9 & 19.5 & 24.5 & 24.3 & 26.8 & 23.9 \\
Gaussian blur & 23.1 & 23.3 & 26.0 & 22.3 & 24.8 & 24.7 & 27.3 & 24.0 \\
Inpainting & 19.0 & 18.6 & 18.1 & 18.4 & 25.5 & 25.2 & 27.9 & 24.7 \\
No & - & - & - & - & \graycell{25.8} & \graycell{25.8} & \graycell{28.5} & \graycell{25.2} \\
\bottomrule
\end{tabular}
\caption{
Results for downstream tasks with NeRV-Enc.}
\label{tab:restore}
\end{table}

\subsection{Weight quantization for efficient storage.}
In this quantization procedure, each element of a vector $\mu$ is mapped to the nearest integer using the linear transformation defined by the formula:
\begin{equation}
  \begin{aligned}
    \mu_i = \text{Round}\left(\frac{\mu_i - \mu_\text{min}}{\text{scale}}\right) &* \text{scale} + \mu_\text{min} , 
    \text{where } \\
    \text{scale} = & \frac{\mu_\text{max} - \mu_\text{min}}{2^{b} - 1} ,
    \label{equa:quant}
  \end{aligned}
\end{equation}
Here, $\mu_i$ represents a vector element, \textit{Round} is a rounding function, $b$ is the quantization bit length, $\mu_\text{max}$ and $\mu_\text{min}$ are the maximum and minimum values of vector $\mu$, and 'scale' is the scaling factor. Additionally, we use Huffman encoding to further reduce the disk storage.

\textit{}{Results for Model Quantization}
We extend our analysis on model quantization in \cref{tab:supple-weight_token_ablation}, assessing performance on three datasets: K400, Something-V2, and UCF-101.

\begin{table}[h!]
\centering
\begin{tabular}{c|ccc|ccc}
\toprule
\multirow{2}{*}{Bits} &  \multicolumn{3}{c}{PSNR $\uparrow$ } & \multicolumn{3}{c}{SSIM $\uparrow$ } \\
 &  K400 & STH-V2 & UCF101 &  K400 & STH-V2 & UCF101 \\
 \midrule
32 & 28.4 & 31.6 & 28.1 & 0.808 & 0.862 & 0.817 \\
\midrule
8 & 28.4 & 31.5 & 28.1 & 0.808 & 0.861 & 0.816 \\
7 & 28.3 & 31.5 & 28 & 0.807 & 0.86 & 0.815 \\
6 & 28.1 & 31.2 & 27.9 & 0.802 & 0.855 & 0.811 \\
5 & 27.5 & 30.2 & 27.3 & 0.784 & 0.836 & 0.794 \\
4 & 25.6 & 27.7 & 25.6 & 0.712 & 0.759 & 0.725 \\
\bottomrule
\end{tabular}
\caption{Ablation study on \textbf{model quantization}.}
\label{tab:supple-weight_token_ablation}
\end{table}


\subsection{Implementation details}

\noindent\textbf{Video Encoding.}
We firstly provide training details of NeRV-Enc below.
\begin{itemize}[nosep, leftmargin=*]
    \item NeRV-Enc:
    \begin{itemize}[nosep, leftmargin=*]
    \item Video: 8 frames, frame stride evenly sample from the whole video, 256$\times$256 resolution. 
    \item Batch size: 32
    \item Patch size: 64
    \item Position embedding dimension for NeRV: 16
    \item Activation layer in NeRV: GeLU
    \item Kernel size for convolution layers in NeRV: 1, 3, 3, 3
    \item Upscale factor for NeRV blocks: 4, 4, 4, 4
    \item Token number for NeRV layers: 4, 128, 64, 0
    \item Token dimensions for NeRV layers: 256, 144, 288, 0
    \item Model dimension and feed-forward dimension for transformer encoder layers: 720 and 2800 for NeRV-Enc of 47.6M, 1600 and 6400 for NeRV-Enc of larger NeRV-Enc (251M)
    \item Dropout ratio in transformer encoder layers: 0 for default training, 0.15 for larger NeRV-Enc and long training 
    \item Optimizer: AdamW
    \item Learning rate: 0.0001
    \end{itemize}
\end{itemize}

\vspace{2pt}
\noindent\textbf{Video Decoding.}
To assess the decoding speed of NeRV-Dec, H.264, RAM, and AV1, we employ a PyTorch dataloader to facilitate parallel decoding. We initially stack the NeRV model weights before inputting them into NeRV-Dec. Regarding video compression, we utilize the `torchvision.io.write\_video' function to store videos, applying various CRF (Constant Rate Factor) settings. For video loading, we experiment with two backends: `decord' and `torchvision.io.read\_video', selecting the one that offers superior performance for H.264 and AV1.

\section{Hurdles When Converting MLP to CNN}
We address the difficulties encountered when converting MLP to CNN for NeRV-Enc, as detailed in \cref{tab:mlp2cnn}. The final model weights $\theta'$ result from the element-wise multiplication of $\hat{\theta}' \in \mathbb{R}^{d_{\text{out}} \times N}$ and video-agnostic weights $\theta_1 \in \mathbb{R}^{C_{\text{out}} \times C_{\text{in}} \times K \times K}$, as illustrated in Figs. 2 and 4 of the main paper.
This  is expressed as
\begin{equation}
    \theta' := \theta_1 * \hat{\theta}'.\text{expand\_as} (\theta_1),
\end{equation}
and is followed by L2 normalization.
\textbf{a)} Omitting the \textbf{\textit{final normalization}} leads to a significant performance drop, from 25.8 to 22.9 for the K400 test PSNR.
\textbf{b)} \textbf{\textit{Convolution initialization of $\theta_1$}} is crucial, as it increases the test PSNR by approximately 1.8.
\textbf{c)} The choice of \textbf{\textit{expansion dimension}} when expanding $\hat{\theta}'$ to match $\theta_1$'s shape is pivotal. 
Expanding along dimension $-2$ (the default choice) produces the best results compared to dimensions -1 and -3.

\begin{table}[h]
\vspace{-0.8em}
\centering
\begin{tabular}{c|cccc}
\textbf{Method} & \textbf{Train} & \textbf{K400} & \textbf{SthV2} & \textbf{UCF101} \\
\midrule
\graycell{NeRV-Enc}  &  \graycell{25.8} & \graycell{25.8} & \graycell{28.5} & \graycell{25.2} \\
w/o Normalize $\theta'$  & 23.3 & 22.9 & 25.7 & 22.9   \\
w/o ConvInit $\theta_1$ & 24.1  & 24   & 26.7   & 23.5   \\
Expand $\hat \theta'$ at dim = -1          & 24.6  & 24.3 & 26.9   & 23.8   \\
Expand $\hat \theta'$ at dim = -3          & 24.2  & 23.9 & 26.5   & 23.4   \\
\bottomrule
\end{tabular}
\caption{
Challenges in converting MLPs to ConvNets, showcasing normalized final weights $\theta'$, convolution initialization of $\theta_1$, and expansion dimension for $\hat \theta'$. PSNR$\uparrow$ showed, the higher the better.
}
\label{tab:mlp2cnn}
\vspace{-0.9em}
\end{table}

\end{document}